\documentclass[lettersize,journal]{IEEEtran}
\usepackage{amsmath,amsfonts}
\usepackage{array}
\usepackage[caption=false,font=normalsize,labelfont=sf,textfont=sf]{subfig}
\usepackage{textcomp}
\usepackage{stfloats}
\usepackage{url}
\usepackage{verbatim}
\usepackage{graphicx}
\usepackage{balance}
\usepackage{cite}
\usepackage{tikz}
\usetikzlibrary{shapes.geometric,arrows.meta,positioning,fit,backgrounds}
\title{A Practical Proprioceptive Data Collection Utility for the Unitree Go1}

\author{Jack A. Smitterberg$^{1}$ and Tan Chen$^{2}$\\
  $^{1}$Michigan Technological University\\
  $^{2}$State University of New York at Binghamton}

\begin{document}

\maketitle

\begin{abstract}
The Unitree Go1 is a widely used platform for legged robotics
research, yet collecting proprioceptive data from it is unnecessarily
difficult: the Unitree SDK is sparsely documented, and ROS1 (Robot Operating System), the Go1's
native middleware, is fully deprecated. We present an open-source
graphical utility that enables researchers to reliably record
proprioceptive data from the Go1, including joint states, IMU (Inertial Measurement Unit) readings,
foot force sensor data, and gyroscope measurements, without requiring
ROS expertise. The tool provides a simple GUI (Graphical User Interface) for starting, stopping,
and labeling recordings in real time. The contribution of this paper is the collection utility. We characterize the
acquisition quality of 51 recordings collected with it, including the achieved
sampling rate and repeated samples. As a secondary, deliberately modest check that
the data is useful, we train a terrain classifier on it. The results are weak: a
Temporal Convolutional Network reaches about 38\% accuracy on held-out recordings
across ten terrain classes (10\% would be chance), and a window-level split that
ignores recording boundaries inflates the same measurement to about 99\%. We
report this as a demonstration that the logged signals carry terrain information,
not as a competitive classifier. The utility is designed to lower the barrier to
collecting Go1 proprioceptive data without ROS.
\end{abstract}


\section{Introduction}

The Unitree Go1~\cite{unitree_Go1} is a widely deployed quadrupedal
robot platform used by academic laboratories and hobbyists for legged
locomotion research. The Go1 remains attractive due to its low cost,
variety of sensors, ample onboard compute, and open-source nature. Continued research interest in the platform is evidenced by recent work on benchmarking locomotion controllers~\cite{akki2025benchmarking}, state estimation~\cite{kang2025simultaneous}, and novel sensing modalities~\cite{vanhauwermeiren2024baro}.

Working with the Go1 in practice, however, presents significant
challenges. The official Unitree SDK is insufficiently
documented~\cite{unitree_sdk}, and the Go1's native middleware
is ROS1, which has reached end-of-life~\cite{ros_noetic_eol}. Unitree's ROS1
package recommends Ubuntu 18.04 and ROS Melodic, both of which are
past end-of-life. The combination of an insufficiently documented SDK, unsupported software, and a deprecated middleware creates a steep barrier to entry for researchers who simply wish to collect data from the robot.

Prior work has demonstrated the value of proprioceptive sensing for legged robot tasks including terrain classification~\cite{villemure2025terrain,buchanan2021haptic} and obstacle estimation~\cite{ramesh2025probe},
but to the best of our knowledge no generalized, easy-to-use
proprioceptive data collection framework has been released for this platform.

This paper presents an open-source, GUI-based utility that allows any researcher to collect labeled proprioceptive data from the Unitree Go1 without requiring ROS knowledge or deep familiarity with the Unitree SDK. The tool records joint states, IMU data, gyroscope data, and foot force sensor data, exporting structured CSV files suitable for direct use in machine learning pipelines. The focus of this paper is the utility, not the classifier. To show that the logged data is useful, we collect a multi-terrain dataset and train a Temporal Convolutional Network (TCN) \cite{lea2016temporal} for terrain classification. This is a demonstration only, and its results are weak. We report accuracy on held-out recordings (about 38\% over ten classes, against 10\% chance) next to the much higher window-level accuracy (about 99\%) that results when recording boundaries are ignored, and we characterize the acquisition quality of the logs, including the achieved sampling rate. The utility is released under the MIT License at \url{https://github.com/RoLACLab/go1-proprioceptive-logger}.

\section{Background}

\subsection{The Unitree Go1 Platform}
The Unitree Go1~\cite{unitree_Go1} is a quadrupedal robot weighing 12~kg with a payload capacity of roughly 3 to 5~kg. It has 12 degrees of freedom (three per leg: body, thigh, and shank joints) driven by precision motors producing up to 35.55~N$\cdot$m peak torque at the knee. The EDU version used in this work is equipped with foot-end force sensors on each leg, onboard IMUs, and onboard Jetson compute modules, providing 16 CPU cores and a 384-core GPU capable of 1.5~TFLOPS. The robot reaches a maximum speed of 3.7~m/s and exposes all sensor data through the Unitree Legged SDK via a high-level API over wired Ethernet or Wi-Fi. 

\subsection{Proprioceptive Sensing in Legged Robots}
Proprioceptive sensing, encompassing joint state, inertial, and contact force information, has proven essential for a range of legged robot tasks~\cite{maravgakis2023contact,yang2023multiimu}. Applications include terrain classification~\cite{villemure2025terrain,buchanan2021haptic},
obstacle estimation~\cite{ramesh2025probe}, ground reaction force estimation~\cite{kang2025simultaneous}. These sensing modalities enable robust locomotion in addition to exteroceptive sensors, which can improve the reliability in adverse conditions such as snow, ice, rain, or low-light environments.

\subsection{Existing Data Collection Approaches}
Existing approaches to proprioceptive data collection on legged robots typically rely on custom ROS-based logging scripts or vendor-specific tooling. For the Go1 specifically, deprecated ROS1 ecosystem and sparse SDK documentation have made consistent data collection difficult~\cite{unitree_sdk}. In addition, certain sensors, like the foot pressure sensors, are not exposed through ROS1. To the best of our knowledge, no open, GUI-accessible proprioceptive data collection utility has previously been published for this platform.

\section{System Design}

\subsection{Architecture Overview}
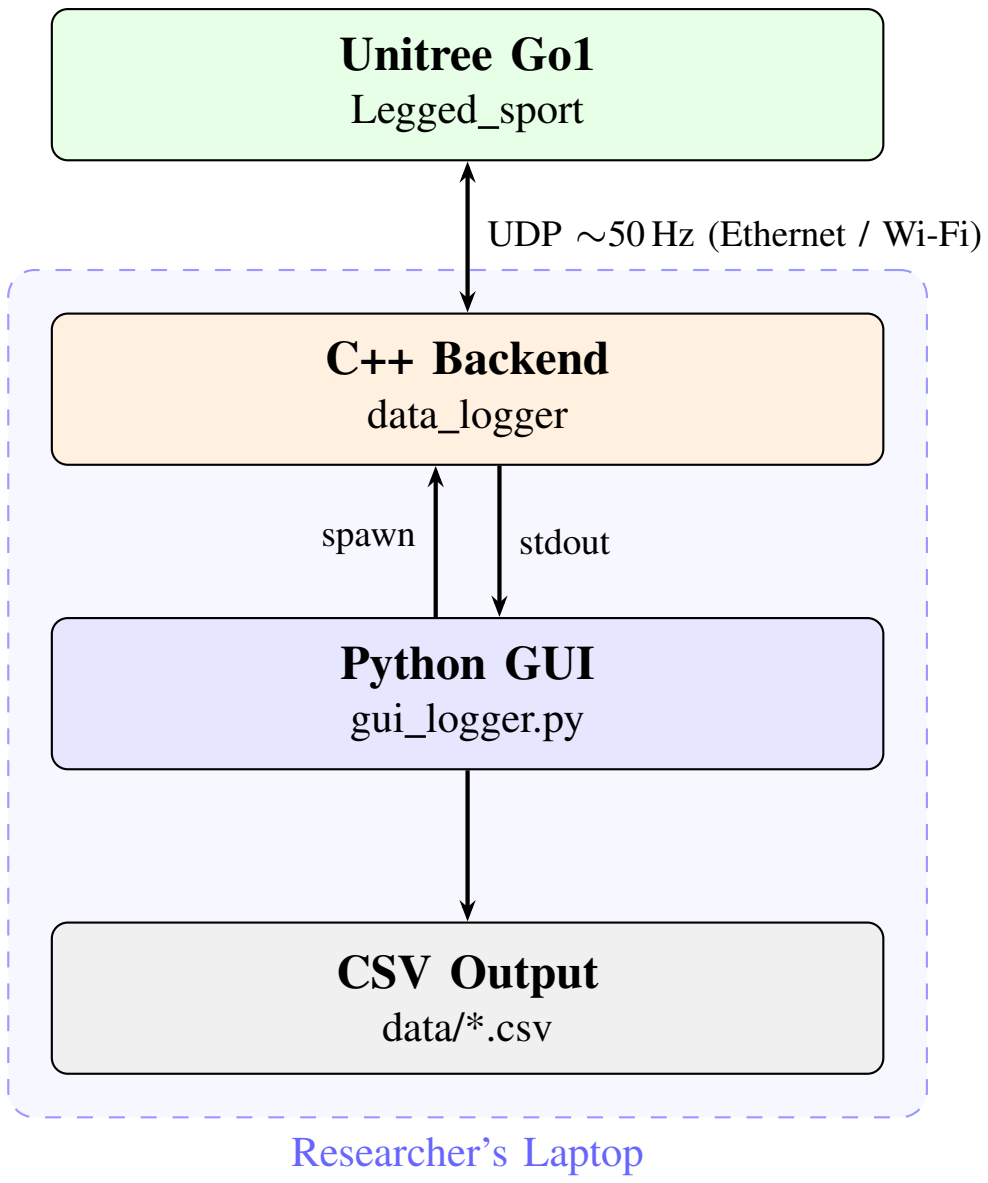
\begin{figure}[t]
  \centering
  \begin{tikzpicture}[
    node distance=0.5cm,
    box/.style={draw, rounded corners=3pt, minimum width=5.5cm, minimum height=1.0cm,
                align=center, font=\small},
    arr/.style={-{Stealth[length=4pt]}, thick},
    darr/.style={{Stealth[length=4pt]}-{Stealth[length=4pt]}, thick},
    lbl/.style={font=\scriptsize}
  ]
    \node[box, fill=green!10] (Go1)
          {\textbf{Unitree Go1}\\{\footnotesize Legged\_sport}};
    \node[box, fill=orange!12, below=1.0cm of Go1] (cpp)
          {\textbf{C++ Backend}\\{\footnotesize data\_logger}};
    \node[box, fill=blue!10,   below=1.0cm of cpp] (gui)
          {\textbf{Python GUI}\\{\footnotesize gui\_logger.py}};
    \node[box, fill=gray!12,   below=1.0cm of gui] (csv)
          {\textbf{CSV Output}\\{\footnotesize data/*.csv}};

    \draw[darr] (Go1) -- (cpp)
        node[lbl, right, midway] {UDP $\sim$50\,Hz (Ethernet / Wi-Fi)};
    \draw[arr] ([xshift= 6pt]cpp.south) -- ([xshift= 6pt]gui.north)
        node[lbl, right, midway] {stdout};
    \draw[arr] ([xshift=-6pt]gui.north) -- ([xshift=-6pt]cpp.south)
        node[lbl, left, midway] {spawn};
    \draw[arr] (gui) -- (csv);

    \begin{scope}[on background layer]
      \node[draw=blue!40, dashed, rounded corners=5pt, fill=blue!3,
            fit=(cpp)(gui)(csv), inner sep=8pt,
            label={[font=\footnotesize,blue!60]below:Researcher's Laptop}] {};
    \end{scope}
  \end{tikzpicture}
  \caption{System architecture. A C++ backend communicates with the Unitree SDK over Ethernet or Wi-Fi via UDP at about 50\,Hz. A Python GUI spawns the backend as a subprocess, reads its stdout stream for live sensor data, and writes labeled CSV recordings.}
  \label{fig:architecture}
\end{figure}

The utility runs on a laptop connected to the Go1 via Ethernet (preferred) or Wi-Fi. The system consists of two layers: a C++ backend that interfaces directly with the Unitree Legged SDK to poll sensor data, and a Python graphical user interface that wraps the compiled C++ executable and provides researcher-facing controls. This architecture avoids any dependency on ROS, allowing the tool to run on modern Ubuntu distributions under Python~3. Fig.~\ref{fig:architecture} illustrates the overall system design.

\subsection{Sensor Streams}
The utility records the following proprioceptive data streams:

\begin{itemize}
  \item \textbf{Joint States:} Position, velocity, and estimated output
        torque (as reported by the robot) for all 12 joints.
  \item \textbf{IMU Data:} Linear acceleration and orientation from
        the robot's onboard IMU(s).
  \item \textbf{Gyroscope:} Angular velocity across all three axes.
  \item \textbf{Foot Force Sensors:} Raw readings from the foot airbag
        sensor of each of the four feet.
  \item \textbf{Foot Kinematics:} Foot position and velocity relative to
        the body for each foot, as reported by the robot.
\end{itemize}

Robot state arrives at about 50~Hz in our setup (Section~\ref{sec:quality}); the SDK documentation does not state this rate. Camera data (exteroceptive sensor) is explicitly out of scope for this utility and is available through other utilities, including the Unitree specific app. The sensor selection is modular: researchers may record any subset of the above streams for a given session.

\subsection{Graphical User Interface (GUI)}
\begin{figure}[t]
  \centering
  \includegraphics[width=\columnwidth]{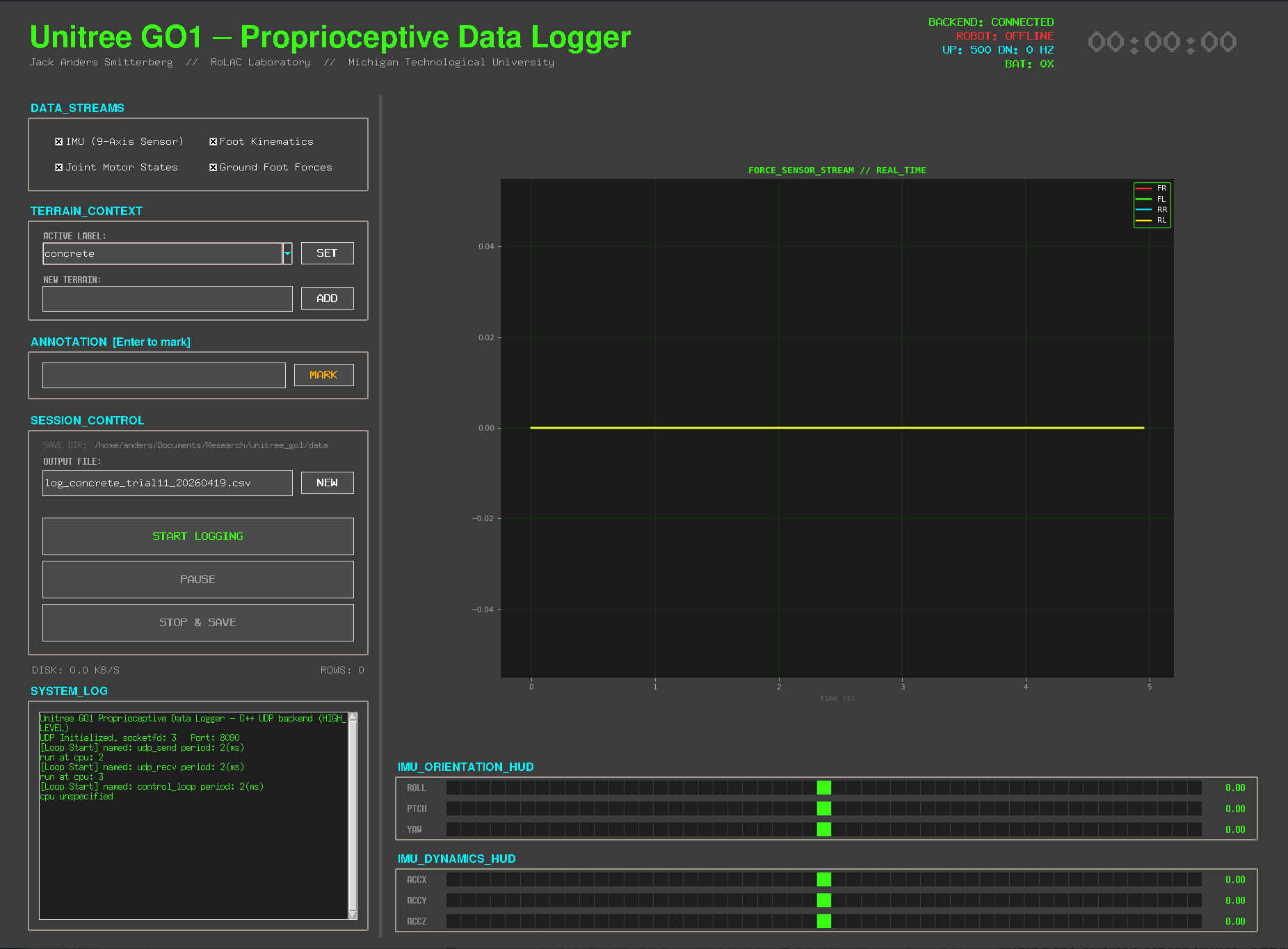}
  \caption{The data collection GUI. Researchers can select sensors, choose or define terrain types, start and stop recordings, add timestamped annotations, and monitor live sensor readouts.}
  \label{fig:gui_screenshot}
\end{figure}

The GUI provides a simple interface for managing recording sessions.
Fig.~\ref{fig:gui_screenshot} shows the main interface.

The typical workflow is as follows. The researcher selects which sensor streams to record and chooses a terrain label from a dropdown menu. Terrain labels are pre-populated with common types but new labels can be created and added at any time. A session filename is auto-populated but remains editable. The researcher presses \textit{Start Logging}, walks the robot over the terrain of interest, and may insert timestamped text annotations at any point during the session, for example to mark gait transitions or flag anomalies. These annotations are written directly into the CSV at the corresponding timestamp. A live readout panel displays the current IMU and foot force sensor values, allowing the researcher to confirm that data is being received correctly. When the session is complete, the researcher presses \textit{Stop \& Save} and the CSV is finalized.

\subsection{Data Output Format}
Each recording session is exported as a CSV file in which each row corresponds to a single timestamped sample containing the selected sensor values. Annotations are embedded as labeled entries at their respective timestamps. This format requires no special parsing library and is directly compatible with standard Python data science libraries such as pandas and NumPy.

Table~\ref{tab:csv_schema} summarizes the 73 sensor channels recorded by the utility, grouped by sensor type. Foot position and velocity are described by the SDK only as relative to the body; it does not document their units or axis directions, so the units listed for them are inferred from the observed value ranges. Joint torque is the robot's estimate, and foot force is a raw sensor reading without documented units.

\begin{table}[t]
  \centering
  \caption{Logged sensor channels and their units. All 12 joints follow
  the order: FR, FL, RR, RL $\times$ hip, thigh, calf.}
  \label{tab:csv_schema}
  \begin{tabular}{llrr}
    \hline
    \textbf{Sensor Group} & \textbf{Fields} & \textbf{Ch.} & \textbf{Units} \\
    \hline
    IMU Acceleration   & \texttt{imu\_acc\_\{x,y,z\}}          &  3 & m/s$^2$ \\
    IMU Gyroscope      & \texttt{imu\_gyro\_\{x,y,z\}}         &  3 & rad/s \\
    IMU Orientation    & \texttt{imu\_\{roll,pitch,yaw\}}       &  3 & rad \\
    Joint Position     & \texttt{*\_\{hip,thigh,calf\}\_q}      & 12 & rad \\
    Joint Velocity     & \texttt{*\_\{hip,thigh,calf\}\_dq}     & 12 & rad/s \\
    Joint Torque (est.) & \texttt{*\_\{hip,thigh,calf\}\_tau}    & 12 & N$\cdot$m \\
    Foot Position      & \texttt{*\_pos\_\{x,y,z\}}             & 12 & m \\
    Foot Velocity      & \texttt{*\_vel\_\{x,y,z\}}             & 12 & m/s \\
    Foot Force         & \texttt{\{FR,FL,RR,RL\}\_force}        &  4 & raw \\
    \hline
    \textbf{Total}     &                                        & \textbf{73} & \\
    \hline
  \end{tabular}
\end{table}

\subsection{Implementation Details}
The C++ backend runs three concurrent \texttt{LoopFunc} threads provided
by the Unitree SDK: a control loop, a UDP (User Datagram Protocol) send loop, and a UDP receive
loop, all scheduled at $dt = 0.002\,s$ (500~Hz). The control loop polls
the latest robot state via \texttt{udp.GetRecv}, but the robot sends
high-level state packets at roughly
50~Hz in our measurements (Section~\ref{sec:quality}), so about nine of every ten polls
return a state that has already been seen. To
avoid logging repeated states, a CSV row is written only when the SDK's
received-packet counter has advanced since the previous poll, so each row
corresponds to one newly received packet and the effective logging rate
equals the packet arrival rate. In addition to the 73 sensor channels, each
row stores the host timestamp of the poll that detected the packet, the
SDK's cumulative received-packet counter, and its cumulative CRC and
packet-loss error counters. Gaps in the packet counter and increases in the
error counters therefore identify dropped or corrupted packets in
post-processing, and the achieved sampling rate and timestamp jitter can be
computed directly from the logs. Timestamps record when a packet was
detected on the host, not when the robot generated it, so they include
network and scheduling latency.

The Python GUI spawns the compiled backend as a subprocess and
communicates with it over standard I/O. Recording commands \texttt{start},
\texttt{stop}, \texttt{pause}, \texttt{resume}, \texttt{terrain},
\texttt{marker}, and \texttt{toggle} are written to the backend's stdin.
The backend emits two structured line types on stdout: \texttt{TELEMETRY}
lines carrying connection status and live sensor readings at 20~Hz, and
\texttt{DATA} lines carrying the full 73-channel sensor vector, one line
per newly received robot state (about 50~Hz), for live consumers such as
the terrain classifier. A dedicated reader thread in the GUI
parses these lines without blocking the main UI thread. CSV rows are
flushed to disk every 500 samples (about $10\,s$ at 50~Hz) to protect
against data loss on unexpected termination.

\section{Terrain Classification Demonstration}

This section is a demonstration, not the main contribution. To show that data collected with this utility can support a downstream machine learning task, we collected a multi-terrain dataset on a physical Unitree Go1 and trained a terrain classifier. The classifier is a simple baseline, and its modest accuracy should be read as a statement about this small dataset, not about the utility. The evidence that matters for the utility is the acquisition quality analysis below.

\subsection{Data Collection Protocol}
Data was collected across ten terrain types: carpet, tiles, concrete, asphalt, snow, deep snow, bricks, ice, stair, and high-friction surface. For each terrain, recordings were made using all three available Go1 gaits: walking, running, and the high-clearance climbing gait. Recordings were made on three days (February 10 and 11, and March 30, 2026), and each terrain was recorded on one to three of these days. All data was collected on the physical robot using the presented utility.

\subsection{Dataset Summary}
Table~\ref{tab:dataset} summarizes the 51 recordings, which total 58.0 minutes of wall-clock time, between 0.4 and 13.9 minutes per terrain. These logs were written by the original logger version, which appended a row on every 2~ms poll of the control loop (1,613,379 rows in total). Most rows repeat the previous robot state, as analyzed in Section~\ref{sec:quality}; only 156,932 rows (9.7\%) are distinct robot states. Class sizes reflect the availability of each terrain type during collection, with concrete and deep snow being the most extensively sampled.

\begin{table}[t]
  \centering
  \caption{Dataset summary. Duration is wall-clock recording time. Distinct
  states are rows whose sensor channels differ from the previous row.}
  \label{tab:dataset}
  \begin{tabular}{lrrrr}
    \hline
    \textbf{Terrain} & \textbf{Files} & \textbf{Dur. (min)} & \textbf{Rows} & \textbf{Distinct} \\
    \hline
    Asphalt       &  1 &  2.4 &    73,150 &   7,303 \\
    Bricks        &  1 &  0.4 &    10,591 &   1,042 \\
    Carpet        &  4 &  4.1 &   123,655 &  12,048 \\
    Concrete      &  8 & 13.9 &   319,991 &  30,133 \\
    Deep Snow     & 12 & 11.6 &   347,068 &  34,607 \\
    High Friction &  5 &  4.8 &   144,247 &  13,465 \\
    Ice           &  6 &  4.9 &   147,545 &  14,712 \\
    Snow          &  7 &  8.5 &   250,718 &  24,221 \\
    Stair         &  2 &  1.5 &    44,281 &   4,410 \\
    Tiles         &  5 &  5.9 &   152,133 &  14,991 \\
    \hline
    \textbf{Total} & \textbf{51} & \textbf{58.0} & \textbf{1,613,379} & \textbf{156,932} \\
    \hline
  \end{tabular}
\end{table}

\subsection{Acquisition Quality of the Collected Logs}\label{sec:quality}
The original log format has no packet counter, so freshness was assessed by comparing each row with the previous one: a row whose 73 sensor channels are all identical to the previous row is a repeated state. By this measure 90.3\% of rows are repeats, and the distinct robot states arrive at a median of 49.9~Hz per recording. The earlier public release of the logger documented a rate of about 60~Hz; the SDK documentation states no rate, and the measured rate is the one we rely on. The interval between consecutive distinct states has a median of 20.0~ms, and 98.8\% of intervals fall between 15 and 25~ms. A further 0.9\% fall between 35 and 50~ms, which is consistent with one missed packet, and 0.2\% are 50~ms or longer, including two stalls of 10.5~s and 49.4~s in two recordings. Ten of the 51 recordings have a distinct-state rate below 48~Hz (between 9.0 and 47.7~Hz); the lowest, concrete trial 7, delivers new states at only 9.0~Hz. We did not exclude any recording from the analysis below.

This analysis has limits. A genuine new packet that is bit-identical to the previous one would be counted as a repeat, although with floating-point IMU and joint channels this is unlikely. The legacy format also cannot distinguish packet loss on the network from stalls on the robot or the logging computer, because the SDK's receive and error counters were not recorded. Timestamps are host-side with 2~ms polling resolution. The current logger release records the packet counter and the CRC and loss counters in every row so that these quantities can be measured directly.

\subsection{Model and Evaluation Protocol}
We trained a Temporal Convolutional Network (TCN)~\cite{bai2018tcn}, an LSTM, and a standard 1D-CNN. The TCN is intentionally shallow so that inference is cheap enough for real-time use. Before training, repeated rows were removed so that each sample is one distinct robot state, and a window of 100 samples therefore spans about 2~s. Channels were normalized with the mean and standard deviation of the training windows only; no hand-crafted feature engineering was applied. Training used Adam (learning rate $10^{-3}$, batch size 64) for 15 epochs. The final epoch is evaluated, with no checkpoint selection and no hyperparameter tuning on test data. Class imbalance was handled with class-balanced sampling and with on-the-fly augmentation (Gaussian jitter, amplitude scaling, and time warping) applied to training data only. Windows were extracted with a stride of 10 samples for training and 50 samples for testing.

We evaluate under two protocols. In the \emph{recording-level} protocol, whole recordings are held out: for each terrain with two or more recordings, about 20\% of the recordings (at least one) are randomly assigned to the test set, so no test window shares a recording with any training window. Asphalt and bricks have a single recording each, so the first 75\% of each recording is used for training and the last 20\% or fewer for testing, with the samples between them (at least 100, one window length) discarded. In the \emph{window-level} protocol, windows from all recordings are randomly assigned 80/20 to training and test sets. Neighboring windows overlap and come from the same recording, so this protocol lets near-duplicate windows appear on both sides of the split. We repeat each protocol over several random splits (five for the TCN recording-level result, three for the TCN window-level result and for the other recording-level results, and one otherwise) and report mean and standard deviation.

\subsection{Results}
Table~\ref{tab:models} reports accuracy under both protocols. Under the recording-level protocol the TCN reaches $37.7 \pm 7.1\%$ over five splits. This is well above the 10\% expected by chance for ten balanced classes and above the 23\% obtained by always predicting the most common test class (concrete), but far below the window-level result of $98.6 \pm 0.3\%$ for the same model on the same data. The 1D-CNN and LSTM show the same gap, and the differences among the three architectures are within the variation across splits, so we do not claim one is better.

\begin{table}[t]
  \centering
  \caption{Test accuracy (\%, mean $\pm$ standard deviation over random
  splits; $n$ is the number of splits) for ten terrain classes.}
  \label{tab:models}
  \begin{tabular}{lcc}
    \hline
    \textbf{Model} & \textbf{Recording-level} & \textbf{Window-level} \\
    \hline
    TCN     & $37.7 \pm 7.1$ ($n{=}5$)  & $98.6 \pm 0.3$ ($n{=}3$) \\
    1D-CNN  & $42.6 \pm 5.1$ ($n{=}3$)  & $99.1$ ($n{=}1$) \\
    LSTM    & $35.5 \pm 12.7$ ($n{=}3$) & $89.5$ ($n{=}1$) \\
    \hline
  \end{tabular}
\end{table}

\begin{figure}[t]
  \centering
  \includegraphics[width=\columnwidth]{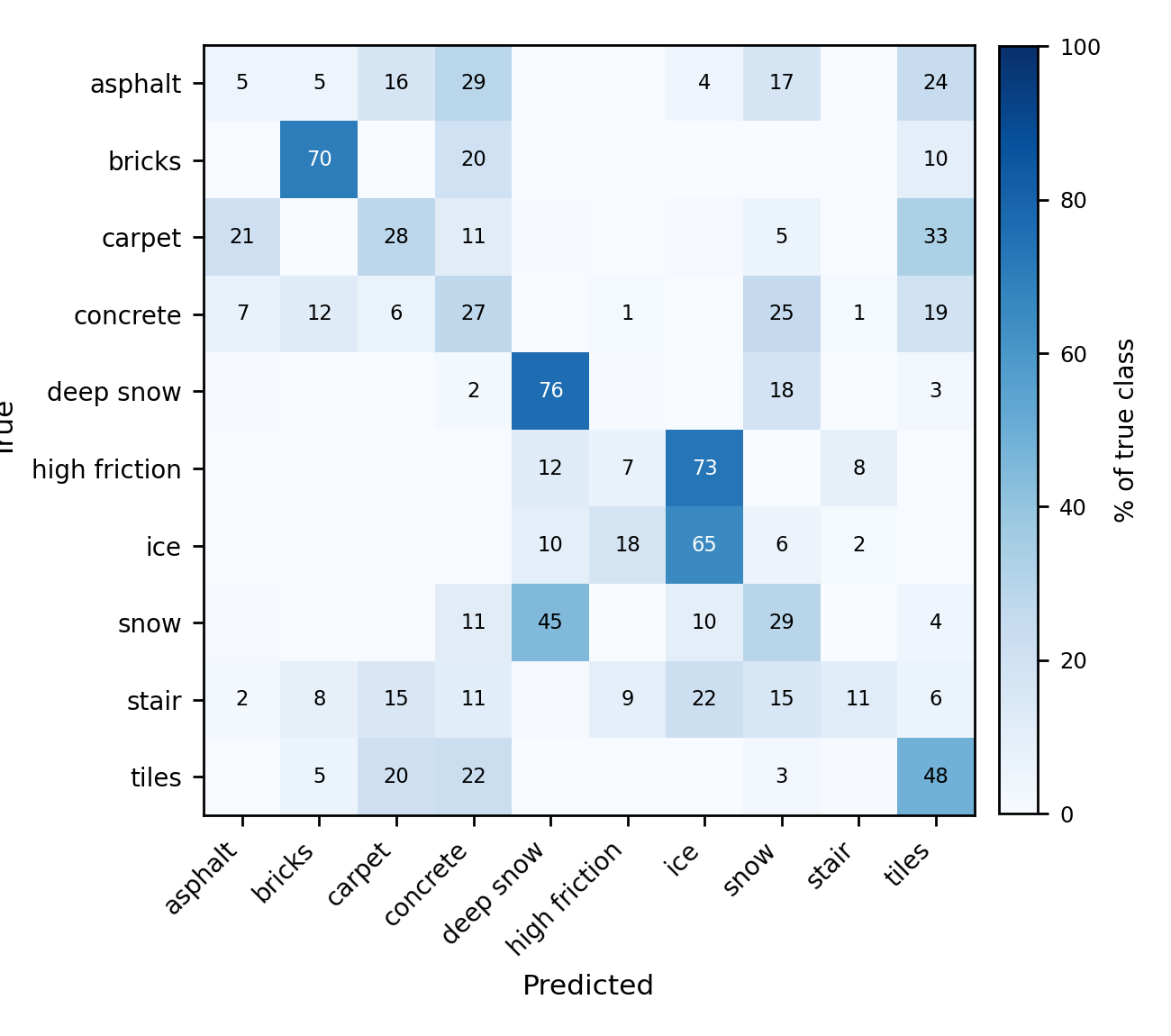}
  \caption{Confusion matrix of the TCN under the recording-level protocol, summed over five random splits. Each row is normalized to the percentage of that true class.}
  \label{fig:confusion_matrix}
\end{figure}

Table~\ref{tab:per_class} and Fig.~\ref{fig:confusion_matrix} give per-class results for the TCN under the recording-level protocol, pooled over the five splits. Deep snow and ice are recognized best, and asphalt, high friction, and stair worst. Bricks has only 10 test windows, and asphalt and bricks are evaluated on a held-out segment of one recording rather than on a held-out recording, so their entries are weak evidence.

\begin{table}[t]
  \centering
  \caption{Per-class recording-level test accuracy of the TCN, pooled over
  five random splits (windows from the same held-out recording can recur
  across splits).}
  \label{tab:per_class}
  \begin{tabular}{lrr}
    \hline
    \textbf{Class} & \textbf{Test windows} & \textbf{Acc.} \\
    \hline
    Asphalt       &  140 &  5.0\% \\
    Bricks        &   10 & 70.0\% \\
    Carpet        &  417 & 27.8\% \\
    Concrete      &  739 & 27.1\% \\
    Deep Snow     &  611 & 76.3\% \\
    High Friction &  209 &  6.7\% \\
    Ice           &  266 & 65.4\% \\
    Snow          &  315 & 29.2\% \\
    Stair         &  235 & 11.5\% \\
    Tiles         &  300 & 48.3\% \\
    \hline
    \textbf{Overall} & \textbf{3,242} & \textbf{38.5\%} \\
    \hline
  \end{tabular}
\end{table}

We did not isolate why held-out recordings are so much harder. Windows from one recording share session-specific conditions such as gait, speed, and battery state, which a model can exploit when they appear on both sides of a split. Terrain is also partly confounded with collection day and gait, because each terrain was recorded on only one to three days. These results show that the logged signals contain terrain information. They do not show that this dataset supports reliable terrain classification on unseen recordings, and we do not present the classifier as competitive.

\subsection{Dataset and Code Availability}
The 51 recordings analyzed here are not currently distributed with the repositories. The classifier repository contains the evaluation code (\texttt{evaluate.py}) that produces the results above, which removes repeated rows from legacy logs before windowing; the logger repository contains a script that reports acquisition quality for any log and a tool that removes repeated rows from legacy logs. Each log file follows the naming convention
\texttt{log\_\{terrain\}\_trial\{n\}\_\{date\}.csv} and is directly
compatible with standard Python data science libraries such as pandas
and NumPy.

\subsection{Real-Time Deployment}
To demonstrate the full pipeline from data collection to on-robot
inference, the companion training utility supports real-time terrain
classification on the physical Go1. The live deployment mode reuses the
same C++ backend as the data logger: the GUI loads a saved checkpoint
comprising model weights, a fitted \texttt{StandardScaler}, a terrain
label map, and architecture metadata, then spawns the backend and reads
its continuous \texttt{DATA} stream. A sliding window buffer accumulates
incoming samples, and once the buffer is full, inference runs every ten new
samples. Because \texttt{DATA} lines are emitted once per newly received state, a live window of 100 samples covers the same roughly 2~s that training windows do. We have not evaluated live prediction accuracy on the robot. The predicted terrain class, its softmax confidence score, and
per-class probability bars are displayed in real time. This closes the
loop between collection and deployment without requiring any additional
software beyond the companion repository~\cite{go1_classifier}.

\section{Discussion}\label{sec:discussion}

We measured a state rate of roughly 50~Hz in the high-level SDK mode; the SDK documentation does not state this rate, and we did not test other robot configurations. Future work may explore low-level SDK access to increase the acquisition rate.

\textbf{Limitations.} The validation in this paper is a demonstration, not a full characterization of the data. First, the recording-level test sets are small (about ten held-out recordings per split), the held-out recordings are chosen at random, terrain is confounded with collection day and gait, and two terrains are evaluated on a time segment of a single recording; the resulting accuracy varies widely across splits. Second, the recordings predate the packet and error counters, so packet loss could not be measured, and we kept every recording, including one whose robot state updates at only 9~Hz. Third, we do not compare the workflow quantitatively against ROS1 logging, raw SDK scripts, or other data collection practices, so the claim that the utility lowers the barrier to Go1 research rests on its design rather than on a measured comparison. Finally, timestamps are host-side and do not account for latency between the robot and the logging computer.

The utility is intentionally scoped to proprioceptive sensing. Camera and depth sensor integration is left for future work.

Although the current implementation targets the Go1, the architecture
is readily adaptable to newer Unitree platforms. The Go2 and G1 each
ship with distinct SDKs from the one used here; migrating to either
platform requires retargeting the C++ backend against the respective
SDK, while the Python GUI layer requires no platform-specific
modification. The high-level access patterns across these
platforms are analogous, so the adaptation effort is expected to be
minor. Labs working with the Go2 or G1 should therefore be able to
extend this utility with minimal effort.

\section{Conclusion}

We have presented an open-source, GUI-based proprioceptive data collection utility for the Unitree Go1. The tool interfaces directly with the Unitree SDK via a C++ backend, eliminating any dependency on ROS, and exposes a Python GUI that allows any researcher to record joint states, IMU data, gyroscope data, and foot force sensor data with minimal setup. Data is exported as structured CSV files ready for use in machine learning pipelines. We characterized the acquisition quality of 51 recordings collected with the utility, finding a distinct-state rate of about 50~Hz and that most rows in the original log format were repeats, which the current release avoids by logging only newly received packets. A TCN terrain classifier reaches about 38\% accuracy on held-out recordings across ten terrain classes, against about 99\% under a window-level split, so the logged signals carry terrain information, but cross-recording generalization on this small dataset is limited. The classifier is a demonstration of the data's value, not the contribution of this work. The utility is available at \url{https://github.com/RoLACLab/go1-proprioceptive-logger} under the MIT License, and the evaluation code is available at
\url{https://github.com/RoLACLab/go1-proprioceptive-classifier}.

\section*{Acknowledgments}

The authors gratefully acknowledge Shivayogi Akki for substantial assistance with data collection on the physical robot. Generative artificial intelligence was utilized to assist with the development and debugging of this code, and preparation of this manuscript.

\bibliographystyle{IEEEtran}
\bibliography{references}

\end{document}